%% file: root.tex
\documentclass[letterpaper, 10 pt, conference]{ieeeconf}  % Comment this line out if you need a4paper

\IEEEoverridecommandlockouts                              % This command is only needed if 
\usepackage{amsthm,amsmath,amssymb}
\usepackage[ruled,linesnumbered]{algorithm2e}
\usepackage{algorithmic}
\usepackage{mathrsfs}
\usepackage{siunitx}
\usepackage{color}
\usepackage{pifont}
\usepackage{cite}
\usepackage{multirow}
\usepackage{url}
\usepackage{xcolor}
\usepackage{xspace}
\usepackage{subfigure} 
\usepackage{graphicx}
\usepackage{comment}

\newcommand{\pcr}{MT-PCR\xspace}
\newcommand{\etal}{\textit{et al.}\@}
\makeatletter
\let\NAT@parse\undefined
\makeatother
\usepackage[colorlinks,
            urlcolor=blue,
            linkcolor=blue,       %%修改此处为你想要的颜色
            anchorcolor=blue,  %%修改此处为你想要的颜色
            citecolor=green        %%修改此处为你想要的颜色，例如修改blue为red
            ]{hyperref}

\title{\LARGE \bf
MT-PCR: Leveraging Modality Transformation for Large-Scale Point Cloud Registration with Limited Overlap
}

\author{Yilong Wu$^{1}$, Yifan Duan$^{1}$, Yuxi Chen$^{2}$, Xinran Zhang$^{1}$, Yedong Shen$^{1}$\\ Jianmin Ji$^{1}$, Yanyong Zhang$^{3}$*~\IEEEmembership{Fellow,~IEEE} and Lu Zhang$^{4}$*% <-this % stops a space
\thanks{* The corresponding author.}
\thanks{$^{1}$School of Computer Science and Technology, University of Science and Technology of China, Hefei, 230026, China
{\tt\small \{elonwu, dyf0202, zxrr, sydong2002\}@mail.ustc.edu.cn, jianmin@ustc.edu.cn}.}%
\thanks{$^{2}$Department of Statistics and Finance, School of Management, University of Science and Technology of China, Hefei, 230026, China
{\tt\small cyx0108@mail.ustc.edu.cn}.}%
\thanks{$^{3}$School of Artificial Intelligence and Data Science, University of Science and Technology of China, Hefei, 230026, China
{\tt\small yanyongz@ustc.edu.cn}.}%
\thanks{$^{4}$Institute of Artificial Intelligence, Hefei Comprehensive National Science Center, Hefei, 230088, China
{\tt\small luzha@ustc.edu.cn}.}%
}

\begin{document}

\maketitle
\thispagestyle{empty}
\pagestyle{empty}

%%%%%%%%%%%%%%%%%%%%%%%%%%%%%%%%%%%%%%%%%%%%%%%%%%%%%%%%%%%%%%%%%%%%%%%%%%%%%%%%
\begin{abstract}

Large-scale scene point cloud registration with limited overlap is a challenging task due to computational load and constrained data acquisition. To tackle these issues, we propose a point cloud registration method, \pcr, based on Modality Transformation. \pcr leverages a Bird’s Eye View (BEV) capturing the maximal overlap information to improve the accuracy and utilizes images to provide complementary spatial features. Specifically, \pcr converts 3D point clouds to BEV images and estimates correspondence by 2D image keypoints extraction and matching. Subsequently, the 2D correspondence estimates are then transformed back to 3D point clouds using inverse mapping.  We have applied \pcr to Terrestrial Laser Scanning (TLS) and Aerial Laser Scanning (ALS) point cloud registration on the GrAco dataset, involving 8 low-overlap, square-kilometer scale registration scenarios. Experiments and comparisons with commonly used methods demonstrate that \pcr can achieve superior accuracy and robustness in large-scale scenes with limited overlap. 
%The code will be open sourced on \url{https://github.com/EloonWu/MT-PCR}.

\end{abstract}

\input{chapters/introduction}
\input{chapters/relatedworks}
\input{chapters/method}
\input{chapters/experiment}
\input{chapters/conclusion}
\input{chapters/acknowledgments}

\begin{comment}
\section*{APPENDIX}

Appendixes should appear before the acknowledgment.

\section*{ACKNOWLEDGMENT}

The preferred spelling of the word ÒacknowledgmentÓ in America is without an ÒeÓ after the ÒgÓ. Avoid the stilted expression, ÒOne of us (R. B. G.) thanks . . .Ó  Instead, try ÒR. B. G. thanksÓ. Put sponsor acknowledgments in the unnumbered footnote on the first page.
\end{comment}

%%%%%%%%%%%%%%%%%%%%%%%%%%%%%%%%%%%%%%%%%%%%%%%%%%%%%%%%%%%%%%%%%%%%%%%%%%%%%%%%

\bibliographystyle{IEEEtran}
\bibliography{IEEEabrv, references}

\end{document}

%% file: chapters/introduction.tex
\section{INTRODUCTION}

Point cloud registration has been a fundamental problem in 3D computer vision and robotics\cite{pomerleau2015review,nagy2018real}, which aims to determine a rigid transformation that aligns two point clouds into a unified coordinate system. Over the past few decades, this problem has been addressed through two primary approaches: handcrafted-descriptor-based methods \cite{dong2018hierarchical,ghorbani2022novel} and learning-based methods\cite{lu2021hregnet,yu2023peal}. Some of these methods have proven to be effective in registering highly similar point clouds or small scenes\cite{zeng20173dmatch,ao2021spinnet}.
% \dnote{See the comments below}
% 1 第一段感觉没必要有这么多的引用，一段就21个引用了，这样看起来好像是罗列但都没看的感觉 Done
% 2 第一段的最后一句话，和第二段的总分关系，看起来不是很明显，我感觉挪到第二段里比较好。Done
% 3 除此之外，我觉得把第三段的第一句可以放在第一段的最后，并且换一个语气，主要说明配准都能在哪些方面work的比较好，然后第二段再话锋一转
% 4 第三段直接分析现有的工作为什么解决不了你的问题

\begin{figure}[t]
	\centering
	\includegraphics[width = \linewidth]{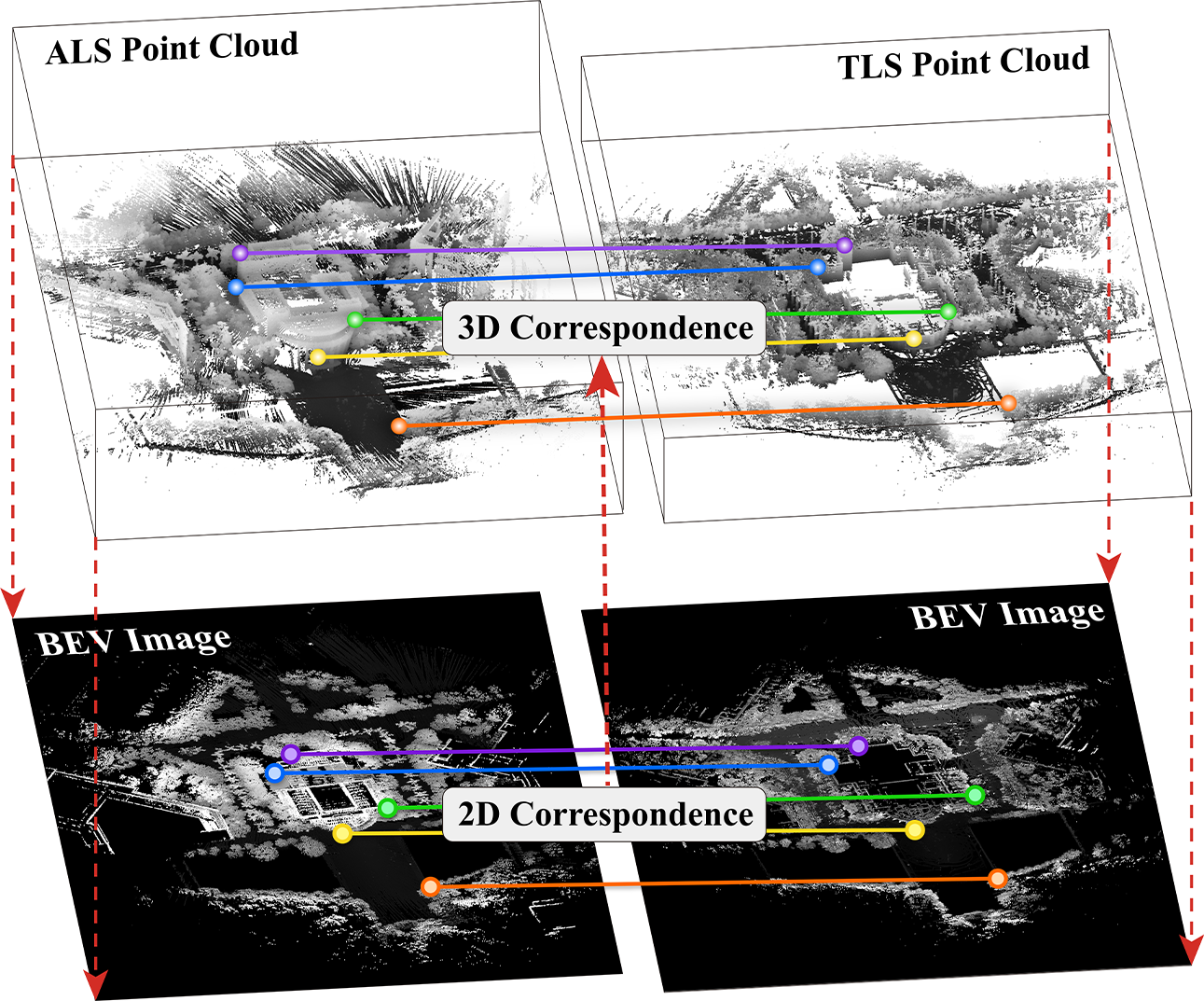}
	\caption{We introduce \pcr, a method for large-scale scene point cloud registration with limited overlap, particularly for TLS-ALS integration. Based on \textbf{M}odality \textbf{T}ransformation, \pcr involves converting 3D point clouds into 2D BEV images, facilitating correspondence estimation through 2D image keypoints extraction and matching.}
	\label{fig:fig1}
\end{figure}

\begin{figure*}[ht]
	\centering
	\includegraphics[width = \linewidth]{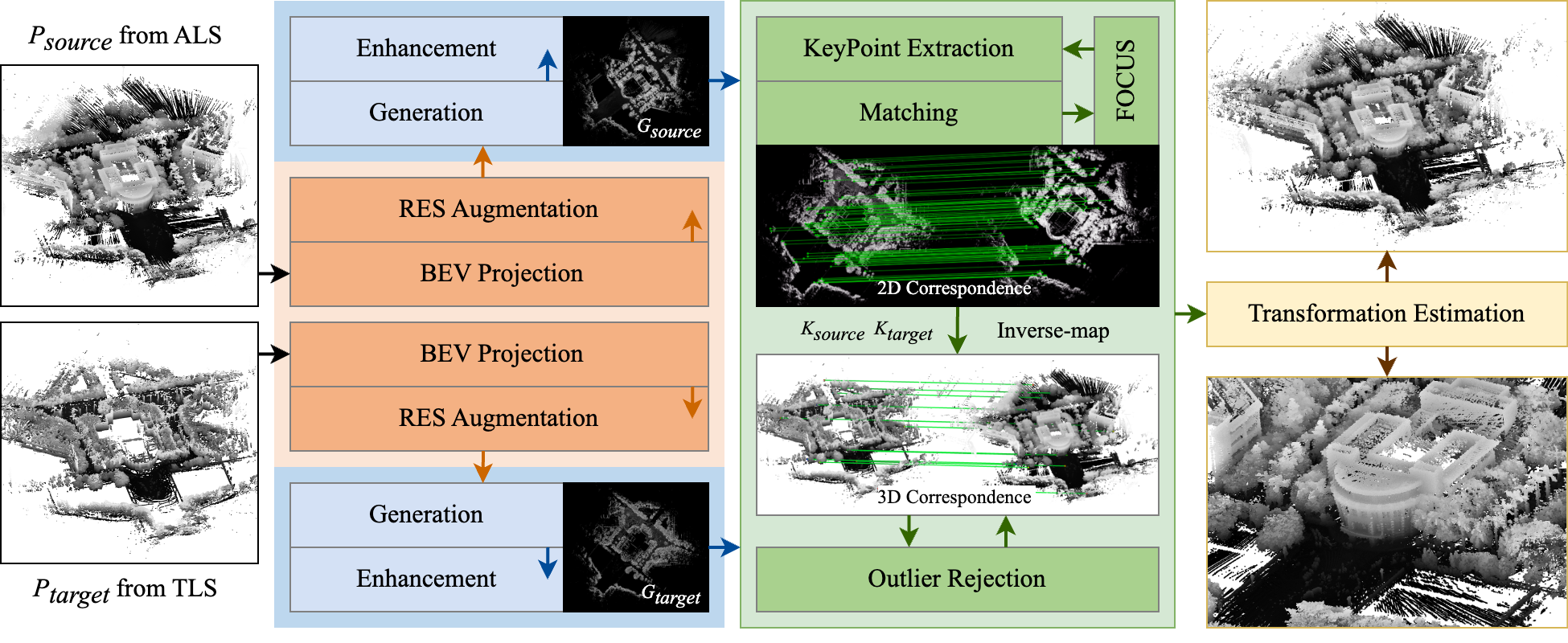}
	\caption{\textbf{Overview of the \pcr pipeline.} The point cloud $\mathcal{P}_{\text{source}},\mathcal{P}_{\text{target}}$ obtained from ALS and TLS is transformed to align with the XOY plane and their resolution is enhanced based on density. Height information is converted to grayscale and further enhanced to emphasize details, resulting in BEV images. Afterward, keypoint extraction and matching are performed using SuperPoint\cite{detone2018superpoint} and LightGlue\cite{lindenberger2023lightglue}, which are repeated in the FOCUS module within overlap regions. 3D correspondences between $\mathcal{P}_{\text{source}}$ and $\mathcal{P}_{\text{target}}$ are established through inverse mapping of 2D correspondences from the BEV image. Finally, the transformation matrix is computed using the SVD algorithm from the refined 3D correspondences.}
	\label{fig:framework}
\end{figure*}

However, the task of registering point clouds in a large-scale scene remains a significant challenge, especially when the overlap between point clouds is limited\cite{prokop2019low,dong2023application}. Large-scale scene point cloud registration suffers from high computational cost associated with processing vast data
% \dnote{datasets?}\xnote{Maybe data or some specific terminology.}\xnote{Notice the space before and after the citation.}
\cite{lei2017fast, xu2019pairwise} and low precision compounded by the presence of noise, missed points and variable local point distribution\cite{xu2019pairwise,li2021automatic}. Furthermore, the limited overlap between point clouds in large-scale scenes is a common issue for practical applications\cite{liu2021coarse,duan2024rotation}, since scanners have to be strategically positioned or moved to guarantee proper overlap between scans, which impose high restrictions on data acquisition process. 
% since obtaining sufficient overlap between point clouds will impose high restrictions on data acquisition process. 
Limited overlap leads to failure in registration due to insufficient information. 

% Current literature and benchmarks mainly focus on small-scale scene point cloud registration \cite{zeng20173dmatch,ao2021spinnet}or only consider pairs of point clouds with $\geq30\%$ overlap for performance measurement\cite{aoki2019pointnetlk,wang2019deep,sarode2019pcrnet}.
Some handcrafted-descriptor-based registration methods have introduced new descriptors with complex structure to address the challenges of large-scale scenes with low overlap~\cite{chen2019plade}, but still constrained by reliance on special structural features and scene complexity. Meanwhile, learning-based registration methods have been specifically optimized for low overlap scenarios but still suffer from excessively high resource consumption in large-scale scenes\cite{huang2021predator}. 
%Additionally, these methods often exhibit limited generalizability, only performing well on specific scenes.

The registration for Terrestrial Laser Scanning(TLS) point cloud and Aerial Laser Scanning(ALS) point cloud represents a typical case of large-scale scene registration with limited overlap\cite{gao2019ground}. Because the data of TLS and ALS point clouds usually consist of millions of points, covering areas of thousands of square meters\cite{zhu2023graco},  
% And the overlap between TLS and ALS point clouds, which can be as minimal as $10\%-30\%$ \dnote{Is there some intuitive understanding of the number 30\%? Why not 40 or 20? Is this a common assumption in registration problems?}\xnote{or some citation and add a statement 'as described in xxx references'.}due to differences in perspective and collection paths, poses a significant challenge for effective registration\cite{he2020ground,jie2023heterogeneous,gao2018accurate}.
and the overlap between TLS and ALS point clouds is limited due to differences in perspectives and collection paths, posing a significant challenge for effective registration\cite{he2020ground,li2021tutorial}.
% While TLS provides high-resolution and high-precision ground detailed information, it is limited in coverage due to view angle constraints. ALS, in contrast, offers broad area coverage with relatively lower resolution and accuracy\cite{gao2018ancient,gao2019ground}. 
% Therefore, the integration of TLS and ALS data is crucial to leveraging their complementary strengths\cite{zhu2020leveraging}, which demands a registration approach development by concentrating on the perspective of large-scale scene registration with limited overlap. \dnote{See the comments below}\xnote{I think it would be clearer to move this sentence to the first instance where TLS and ALS are mentioned and to add another concluding sentence here.}
Therefore, the integration of TLS and ALS data demands a registration approach development by concentrating on the perspective of large-scale scene registration with limited overlap.
% 这一段说实话稍显突兀，看起来也像是背景介绍。一篇文章有了两个背景。感觉如果把背景定在点云配准上的话，一个可能比较好的写法是，“点云配准的背景-挑战-你提出的方法-最后你用TLS-ALS的场景进行算法验证”，把空地这个背景再弱化一下，变成只是数据集。
% 或者就是把空地和前面的背景揉在一起，但这就比较难了，没想好

% In this work, we propose a novel\xnote{delete novel, it says nothing. Do you consider to name your method? If do this you can say, we propose xxx, a method to xxx.} approach to large-scale scene point cloud registration with limited overlap, particularly for TLS-ALS integration. 
In this work, we propose \pcr, a method for large-scale scene point cloud registration with limited overlap, particularly for TLS-ALS integration. 
Based on modality transformation, our proposed method involves converting 3D point clouds into 2D Bird's Eye View (BEV) images\cite{mallot1991inverse}, facilitating correspondence estimation through 2D image keypoints matching. The BEV perspective, which captures extensive overlap information, enhances registration accuracy in scenarios with limited overlap. Utilizing 2D BEV images for keypoints extraction and matching substantially decreases the computational load for large-scale scene tasks involving millions of points. Furthermore, the proposed method overcomes the restrictions of traditional handcrafted-descriptor-based methods in terms of descriptor design and the generalization challenges encountered by learning-based methods, which results in a more effective and robust approach for point cloud registration in large-scale scenes. Additionally, our method incorporates specialized data processing to boost the efficiency of TLS-ALS registration.

% Registration experiments of TLS-ALS point clouds are performed on the GRACO dataset\cite{zhu2023graco}, and a translation error of less than \SI{2}{\meter} and a rotation error of less than \SI{1}{\degree} are achieved in the tens of thousands of square meters level large-scene registration, which outperforms the existing point cloud registration methods. The main contributions of this paper are summarized as follows. 
Registration experiments of TLS-ALS point clouds are performed on the GrAco dataset\cite{zhu2023graco}, and a translation error of less than \SI{2}{\meter} and a rotation error of less than \SI{1}{\degree} are achieved in large scene registration of more than \SI{200000}{\square\meter}
, which outperforms the existing point cloud registration methods. The main contributions of this paper are summarized as follows. 

\begin{itemize}
    % \item Our method achieves accurate registration in limited overlap scenarios by employing a Bird’s Eye View(BEV) perspective, effectively addressing the challenges and costs associated with data acquisition in large-scale scenes.
    % \item Our method leverages 2D image keypoints matching based on modality transformation, significantly lowering the computational burden for large-scale registration tasks.
    % \item Our method enhances registration robustness and generalization, being independent of descriptor design and dataset-specific training. This flexibility allows it to be readily adapted to various complex real-world scenarios. 
    \item We introduce \pcr, incorporating 2D image keypoints matching into point cloud registration based on modality transformation, significantly lowering the computational burden for large-scale registration tasks.
    \item We achieve accurate registration in limited overlap scenarios by employing a BEV perspective, effectively addressing the challenges and costs associated with data acquisition in large-scale scenes.
    \item Through extensive experiments, we demonstrate that our method enhances registration robustness and generalization in the application of TLS-ALS integration, independent of descriptor design and dataset-specific training.
\end{itemize}
% \xnote{The contribution can be written better. These three paragraphs except for the statement of the experiment results aim to describe your contribution, not how good your work is. I mean 'effectively addressing xxx' is not a priority over the description. Of course, it is not absolute, you can present how good it is only necessary and proper. So, just describe it objectively, instead of praising it.}
% The rest of the paper is organized as follows: Section II \xnote{use reference, not hard writing.}will review some related work. Section III provides details of the research theory. Experimental results and evaluation results are discussed in Section IV. Finally, Section V summarizes the paper.\xnote{Is this paragraph necessary? It seems somewhat redundant to me and does not align with the effect you aim to achieve based on the other paper, maybe. If you’d like, it’s okay.}

%% file: chapters/relatedworks.tex
\section{Related Work}

Point cloud registration methods can be divided into two main categories: coarse registration and fine registration. Fine registration algorithms are designed to refine an initial coarse registration, such as Iterative Closest Point (ICP)\cite{besl1992method} and its variants\cite{rueckert1999nonrigid,rusinkiewicz2001efficient,li2008global,segal2009generalized}. 
% Conversely, coarse registration\xnote{according to the above, maybe you should introduce what is coarse registration, and then the difficulty or the differences?} poses greater challenges due to the unknown nature of the initial registration. 
In contrast, coarse registration algorithms deal with point clouds of unknown orientations, making it a more challenging task. 
% Our method pertains to coarse registration in the context of large scenes with limited overlaps. \xnote{this is a section of related work, if you want to describe the differences between ours and existing work, move to the end of the related paragraph, may it will be better.}
% This section will predominantly concentrate on recent advancements in coarse registration, particularly focusing on descriptor-based registration and image-supported registration. \xnote{This sentence is redundant, you can only do this, not other choice.}

\textbf{Descriptor-Based Registration.} The algorithms falling within this category are widely employed for point cloud registration, primarily aimed at devising local salient point features to establish correspondences between two point clouds\cite{gelfand2005robust,qi2017pointnet,elbaz20173d,deng2018ppfnet,choy2019fully}. The typical procedure involves the extraction of keypoints and computation of their descriptors~\cite{rusu2009fast,tombari2010unique}, followed by the establishment of sparse correspondences between the keypoints based on the descriptors. Subsequently, various methodologies have been developed to effectively eliminate false correspondences\cite{wolfson1997geometric,fischler1981random,chen1999ransac,yang2020teaser}.

The registration of point clouds in large-scale scenes with limited overlap has garnered significant attention recently due to its complexity. Chen \etal\cite{chen2019plade} introduced a descriptor founded on high-level structural information (e.g., planes, lines, and their interrelationships) to address the challenge of low overlap in urban scenes. Huang \etal\cite{huang2021predator} presented a deep learning model specifically crafted to facilitate point cloud registration in low-overlap areas, achieved by devising an attention module for overlaps to ascertain the probability of points in the point cloud residing in the overlapping region of the two point clouds. Lu \etal\cite{lu2021hregnet} proposed a methodology for conducting registration on keypoints and descriptors extracted in a hierarchical manner, thereby enabling the registration of large-scale point clouds.

\textbf{Image-Supported Registration.} Recently, a multitude of algorithms have been introduced for registering point clouds using multimodal data\cite{ren2022corri2p,zhang2022pcr}. The primary objective of current multimodal registration algorithms is to enhance the structural information of point clouds through the integration of textural information extracted from images. Yu \etal\cite{yu2023peal} established correspondence in the RGB-D image and generated an overlapping prior of the point cloud, which was then integrated into the transformer to facilitate the registration of point clouds with low overlap. Chen \etal\cite{chen2022imlovenet} employed triple features from the 2D image domain, 3D domain and simulated 3D domain, and fuse them with attention modules.
Similarly, Huang \etal\cite{huang2022imfnet} proposed a multimodal fusion method to construct a descriptor for point cloud registration that considers both structural and textural information extracted from the image. While these methods all necessitate additional sensor information, our proposed method directly generate the image from the point cloud, thus reducing dependence on sensor types.
% \xnote{Make sure you are right, I did not check the above three paragraphs.}

%% file: chapters/method.tex
\section{Methodology}

% For large-scale scene with low-overlap applications such as TLS-ALS point cloud registration, we propose a method based on keypoint matching of Bird's Eye View (BEV) images derived from point clouds. By leveraging the BEV perspective, we can maximize the utilization of point cloud information and overcome the challenge of low overlap. Keypoint extraction and matching based on 2D BEV images significantly reduce the computational demand for large-scale registration tasks involving millions of points. Additionally, our method incorporates specialized data processing to boost the efficiency of TLS-ALS registration. 
% \xnote{It seems to be an overview of this work, but..., seems a little..., um, I have no idea to explain it. Consider shortening it if no space finally. }
Based on modality transformation, \pcr involves converting 3D point clouds into 2D BEV images, facilitating correspondence estimation through 2D image keypoints extraction and matching. 
The process of our method is illustrated in Fig.\ref{fig:framework}.
 % We describe the details of our method below. 

%【改】
% In the following sections, we first introduce the optimization processing of point clouds in the context of TLS-ALS integration scenarios, then detail the generation and enhancement of point cloud BEV images, followed by the estimation of point cloud correspondences, and conclude with the registration algorithm. The process of this method is illustrated in the accompanying figure.
\subsection{Point Cloud Processing}

% 地空点云数据的问题
% 1.无人车和无人机的作业高度不同，采集路线不同，点云重合度低
% 2.采集视角不同，点云的密度，分布和噪声形态有较大差异
% 我们利用以下的点云数据方法针对性的克服数据的复杂和困难。
There are two main issues when registering TLS and ALS point clouds. First, differences in scanning altitudes and routes between unmanned ground vehicles (UGVs) and unmanned aerial vehicles (UAVs) lead to limited overlap of TLS and ALS data on a global scale. Second, the different scanning perspectives and laser scanning technologies of UGVs and UAVs result in variations in point cloud density, distribution, and noise patterns. We employ the following point cloud data processing approaches to overcome the complexity of the data.

\textbf{BEV Projection. }The core of point cloud registration lies in extracting adequate similar features within two point clouds to establish point correspondences. When scanning the same building or structure using both UGA and UAV at different heights, we notice that the point clouds have highly similar contours from a bird's eye view (BEV). %Consequently, when the spatial overlap in point clouds cannot be guaranteed, projecting the point clouds onto the ground from a BEV perspective can yield more similar information. %

To facilitate the subsequent generation of BEV images, we rotate the point cloud such that the ground plane becomes parallel to the XOY plane. Specifically, we employ the RANSAC algorithm to obtain the ground plane of the point clouds, and then calculate the rotation matrix based on the angle between the ground plane and the XOY plane. After this rotation, projecting the point onto the XOY plane is equivalent to projecting onto the ground plane.

%在计算机图像学中，分辨率表示每英寸的像素数量，在本文中还表示每英寸激光点的数量，这直接影响着BEV图像的质量，过低的分辨率会压缩点云细节，过高的分辨率会产生空洞，破坏原有点云的结构。考虑到地面和空中采集点云的密度有较大的差异，我们设计了一种适用于不同点云密度的分辨率
\textbf{Resolution Augmentation. }In computer graphics, resolution denotes the number of pixels per inch, while it also denotes the number of laser points per inch in this study, which significantly affects the quality of BEV images. Excessively low resolution can result in the loss of point cloud details due to compression, while excessively high resolution may create holes that disrupt the original point cloud structure. Given the huge variation in density between TLS and ALS point clouds, we design a resolution parameter $\text{RES}$ that adapts to different point cloud densities. 
\begin{equation}
\label{eq:resolution}
\text{RES} = \frac{N\gamma}{(x_{\text{max}}-x_{\text{min}})(y_{\text{max}}-y_{\text{min}})}
\end{equation}
The resolution for a point cloud $\mathcal{P}$ is calculated by Eq.\ref{eq:resolution}, where $N$ denotes the number of points in $\mathcal{P}$, $\gamma$ is a manually specified hyperparameter, $x_{\text{max}},x_{\text{min}},y_{\text{max}},y_{\text{min}}$ represent the maximum and minimum values in X and Y directions, and the denominator is the area occupied by $\mathcal{P}$. Denote the source point cloud as $\mathcal{P}_{\text{source}}$ and target point cloud as $\mathcal{P}_{\text{target}}$. We process the point clouds by $\mathcal{P}_{\text{source}}'=\mathcal{P}_{\text{source}}\cdot \text{RES}_{\text{source}}$ and $\mathcal{P}_{\text{target}}'=\mathcal{P}_{\text{target}}\cdot \text{RES}_{\text{target}}$. This procedure ensures an increase of point-to-point distances for high density point clouds to clarify details and a decrease of point-to-point distances for low density point clouds to guarantee a continuous contour while reducing holes occurrence or false noise recognition.

\subsection{BEV Image Generation and Enhancement}

Involving full information of millions of points for large-scale scene registration is an infeasible task for existing methods. However, images maintain the same orderliness as point clouds in the X and Y axes while storing information in each pixel, which is an efficient data compression form. Moreover, computer graphics offers advanced methods for image keypoint extraction and matching. Therefore, converting point clouds into images can significantly reduce the processing difficulty associated with massive data in large-scale scene tasks.

\textbf{BEV Image Generation.}
To convert 3D point clouds to BEV images, we first establish a mapping from the point cloud coordinate system to pixel coordinate system. For a point cloud $\mathcal{P}$ with $N$ points, where each point is denoted as $p_k = (x_k, y_k, z_k)$ for $1 \leq k \leq N$, we define that $\mathcal{D}_{ij}=\{ p_k \in \mathcal{P} | \lceil x_k-x_{\text{min}} \rceil = i, \lceil y_k-y_{\text{min}} \rceil = j \}$, which denotes the set of points that project to the pixel at coordinates $(i, j)$. Next we define $H_{ij}$ as the maximum height value of the points in $\mathcal{D}_{ij}$, which can be expressed as:
\begin{equation}
\label{eq:H_matrix}
H_{ij} = 
\begin{cases}
  \max\{ z_k \mid p_k \in \mathcal{D}_{ij} \} & \text{if } \mathcal{D}_{ij} \neq \varnothing, \\
  \text{NaN} & \text{if } \mathcal{D}_{ij} = \varnothing.
\end{cases}
\end{equation}
Using $\mathbf{H}$, we generate the BEV image matrix $\mathbf{G}$, where each element \( G_{ij} \) corresponds to the grayscale value of the pixel at coordinates \( (i, j) \). This can be formulated as:
\begin{equation}
\label{eq:G_matrix}
G_{ij} = 
\begin{cases}
    \lfloor{\frac{H_{ij}-z_{\text{min}}}{z_{\text{max}}-z_{\text{min}}}}\rfloor \times 255 & \text{if } H_{ij} \neq \text{NaN}, \\
  0 & \text{if } H_{ij} = \text{NaN}
\end{cases}
\end{equation}

\textbf{Image Enhancement.} We use image enhancement techniques to improve the accuracy of image keypoint extraction. Two high-pass filters with different strengths are utilized to augment the high-frequency content within the images, thereby improving the distinguishability of structural elements such as buildings, roads, and vegetation. The process can be summarized as follows:
\begin{equation}
\label{eq:image_enhance}
\begin{aligned}
\Tilde{G}_{ij}= (\mathbf{w}_2 & \ast  (\mathbf{w}_1 \ast \mathbf{G}))_{ij}, \\
\mathbf{w}_1 = \left[\begin{array}{ccc}
-2 & -2 & -2 \\
-2 & 32 & -2 \\
-2 & -2 & -2
\end{array}\right] & ,
\mathbf{w}_2 = \left[\begin{array}{ccc}
-1 & -1 & -1 \\
-1 & 10 & -1 \\
-1 & -1 & -1
\end{array} \right]
\end{aligned}
\end{equation}
where we denote $\mathbf{w}_1$ as a convolution kernel for extracting edges, $\mathbf{w}_2$ as a convolution kernel for sharpening details, $\star$ as convolution operation and $\Tilde{\mathbf{G}}$ as the enhanced BEV image matrix\cite{gonzalez2009digital}. 

\subsection{Correspondence Estimation}

\textbf{2D Keypoint Extraction and Matching. }SuperPoint\cite{detone2018superpoint} is a SOTA method among numerous image keypoint extraction methods. %, such as ORB\cite{rublee2011orb}, SIFT\cite{lowe2004distinctive} etc.
Consequently, we utilize SuperPoint to extract keypoints of BEV images and then obtain the 2D point correspondence by the image keypoint matching framework LightGlue\cite{lindenberger2023lightglue}. For the source point cloud and the target point cloud, we denote $\mathcal{K}_{\text{source}}$ and $\mathcal{K}_{\text{target}}$ as the set of matched keypoints from 2D BEV images, respectively. 

\begin{figure}[t]
	\centering 
	\subfigure[Before FOCUS] {\includegraphics[width = 0.48\linewidth]{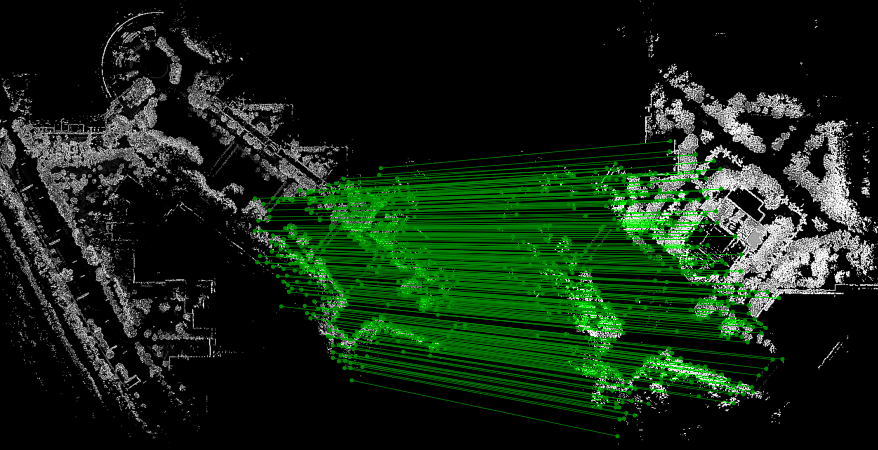}}
	\subfigure[After FOCUS] {\includegraphics[width = 0.48\linewidth]{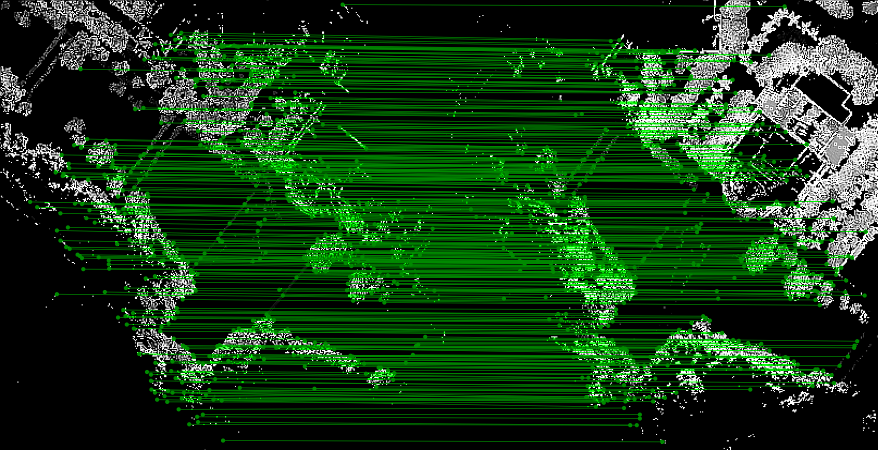}}
	\caption{The comparison of before and after using the FOCUS module. By performing 2D keypoint extraction and matching again on the neighborhood of matched points, the FOCUS module increases the number of matched points and enhances the accuracy of correspondence estimation.}
	%Examples of \metric and \core.}
	\label{fig:focus}
\end{figure}

\textbf{FOCUS Module.} For excessively low overlap between two point clouds, we design a FOCUS module to improve registration accuracy. If the proportion of overlap areas in point clouds is lower than a predefined threshold $\theta$, keypoint extraction and matching are performed again on the neighborhood of the matched points derived from the first matching. Fig.\ref{fig:focus} illustrates the effect of the FOCUS module.

\textbf{3D Correspondence Construction. } 
Based on the 2D keypoint matching results, we remap the keypoints and their correspondence to 3D point clouds. For a matched keypoint in $\mathcal{K}_{\text{source}}$ or $\mathcal{K}_{\text{target}}$ with coordinate $(i,j)$ in 2D images, we denote $(X,Y,Z)$ as its coordinate in the original 3D point cloud, calculated by Eq.\ref{eq:inverse_coordinate}. 
\begin{equation}
\label{eq:inverse_coordinate}
\begin{cases}
    X=(i+x_{\text{min}})/\text{RES}\\
    Y=(i+y_{\text{min}})/\text{RES}\\
    Z=[G_{ij}(z_{\text{max}}-z_{\text{min}})/255+z_{\text{min}}]/\text{RES}
\end{cases}
\end{equation}

\subsection{Transformation estimation}
We employ SVD\cite{golub1971singular} to estimate the transformation matrix between point clouds. To enhance the precision of the estimation, we adopt an iterative optimization strategy. Specifically, by iteratively executing the SVD process, outlier correspondences can be effectively rejected. After the initial estimation of the transformation, we input it as an initial value into the ICP\cite{besl1992method} algorithm to achieve a more refined point cloud registration. 

%% file: chapters/experiment.tex
\section{Experiments}

\begin{figure*}[ht]
	\centering
        \subfigure[Initial Position] {\includegraphics[width = 0.19\linewidth]{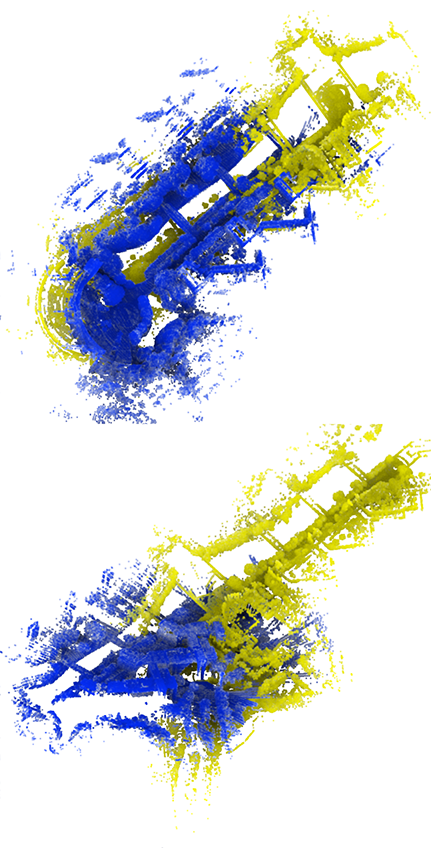}\label{fig:result_a}}
        \subfigure[GT] {\includegraphics[width = 0.19\linewidth]{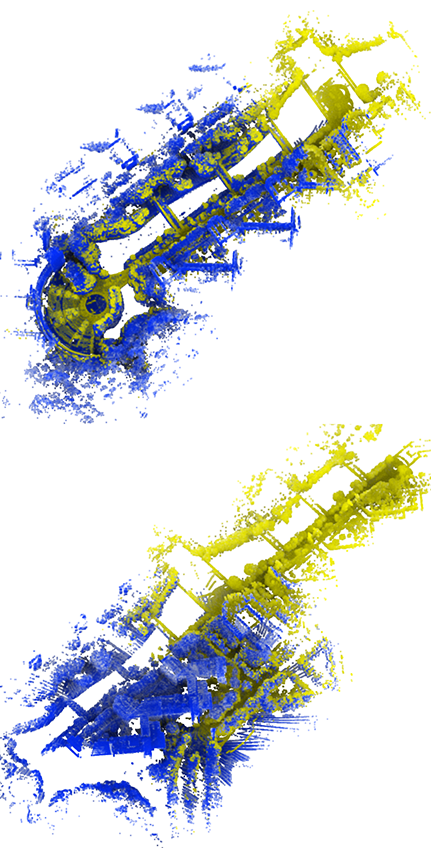}\label{fig:result_b}}
        \subfigure[\pcr(Ours)] {\includegraphics[width = 0.19\linewidth]{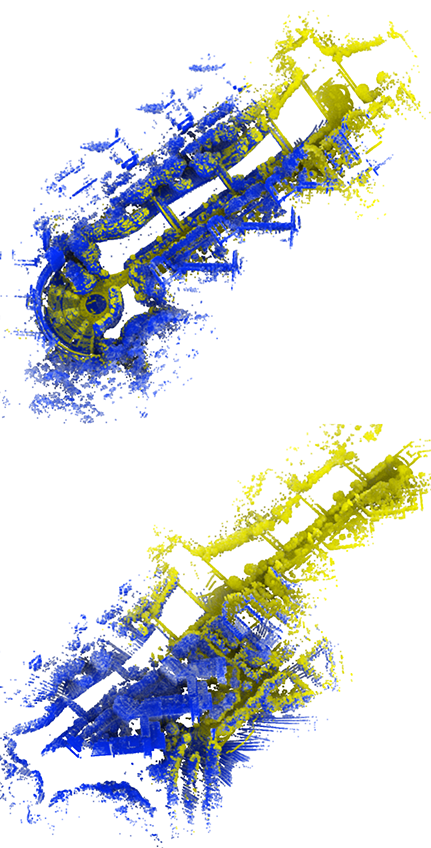}\label{fig:result_c}}
        \subfigure[FPFH+TEASER] {\includegraphics[width = 0.19\linewidth]{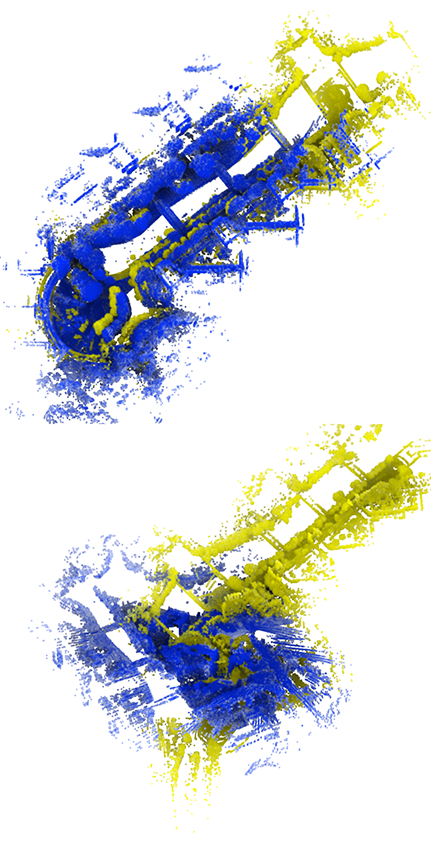}\label{fig:result_d}}
        \subfigure[GeoTransformer] {\includegraphics[width = 0.19\linewidth]{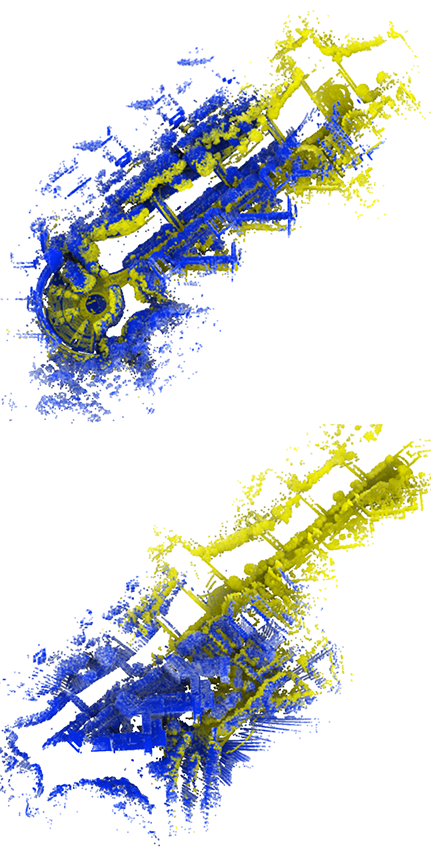}\label{fig:result_e}}
	\caption{Registration results of the \pcr, FPFH, and GeoTransformer on the A03-G03 and A04-G06 combinations. Yellow and blue represent point clouds obtained from TLS and ALS, respectively.}
	%Examples of \metric and \core.}
	\label{fig:result}
\end{figure*}

\subsection{Dataset}
We utilize the GrAco dataset\cite{zhu2023graco} to evaluate the efficacy of \pcr. GrAco represents a high-quality multimodal dataset to examine collaborative simultaneous localization and mapping (SLAM) algorithms for ground and aerial robots. GrAco contains six ground (TLS) sequences and eight aerial (ALS) sequences. Using ``G'' for ground sequences and ``A'' for aerial sequences, our experiments focus on eight A-G, three G-G, and three A-A combinations for source and target point clouds. Table.\ref{table:dataset} provides scale and overlap information of several typical combinations in our experiments, showing that multisource combinations (A-G) tend to have lower overlap ratios compared to unisource combinations (G-G or A-A). 

% This dataset encompasses both ground and aerial perspectives in expansive urban environments and incorporates meticulously calibrated LiDAR, camera, and IMU data, accompanied by centimeter-level ground-truth localization obtained through RTK GNSS. In total, the GrAco dataset comprises six ground-based sequences and eight aerial sequences, featuring intricate scenarios such as buildings, squares, passages, and rooftops.
\input{table/dataset}

\subsection{Evaluation Metric}
\input{table/rot_trans_error}
The performance of \pcr is evaluated using the following three metrics:

\textbf{Rotational and Translation Error. }The residual transformation $\Delta T_{s, t}$ of transformation $T_{s,t}$ from source point cloud $\mathcal{P}_{\text{source}}$ to target point cloud $\mathcal{P}_{\text{target}}$ is defined as follows:
\begin{equation}
\label{eq:error_tsfm}
\Delta T_{s, t}=T_{s, t}\left(T_{s, t}^{G}\right)^{-1}=\left[\begin{array}{cc}
\Delta R_{s, t} & \Delta t_{s, t} \\
0 & 1
\end{array}\right]
\end{equation}
where $T_{s, t}^{G}$ is the ground-truth transformation from $\mathcal{P}_{\text{source}}$ to $\mathcal{P}_{\text{target}}$. Then, the rotation error $e_r$ and translation error $e_t$ are calculated based on the rotational component $\Delta R_{s, t}$ and the translation component $\Delta t_{s, t}$ as follows:
\begin{equation}
\label{eq:rot_trans_error}
\left\{\begin{array}{c}
e_{s, t}^{r}=\arccos \left(\frac{\mathrm{tr}\left(\Delta R_{s, t}\right)-1}{2}\right) \\
e_{s, t}^{t}=\left\|\Delta t_{s, t}\right\|
\end{array}\right.
\end{equation}

\textbf{Root Mean Square Distance(RMSD). }RMSD is the mean distance computed between the transformed source point cloud and the same point cloud in its ground-truth position.

\textbf{Successful Registration Rate(SRR). }Upon establishing predefined rotation and translation error thresholds($\sigma_{r}$ and $\sigma_{t}$), a registration is deemed successful when both the rotation error and translation error are lower than their respective thresholds. SRR is the ratio of the number of successful registrations to the total number of registrations. In this paper, the $\sigma_{r}$ and $\sigma_{t}$ are set as \SI{5}{\degree} and \SI{2}{\meter} according to the application scene.

\subsection{Implementation Details}
% We screened eight sets of ground-aerial sequence combinations with overlapping trajectories from the GrAco dataset to test the performance of the proposed method and compare the performance with other point cloud registration methods. Six sets of ground-ground sequence combinations and six sets of aerial-aerial sequence combinations experiments are also designed to demonstrate the robustness and outstanding performance of the proposed method for point cloud registration processing in large outdoor scenes. 
We conducted an evaluation of eight sets of A-G combinations with overlapping trajectories from the GrAco dataset to assess the efficacy of \pcr. 
% Furthermore, we compared its performance with other point cloud registration methods. 
Additionally, we formulated experiments involving three G-G and three A-A combinations to show the robustness and exceptional performance of \pcr in large outdoor scenes with limited overlap.

% The initial target point cloud $P_t^i$ and source point cloud $P_s^i$ are obtained by superimposing the LiDAR raw data, and each frame of LiDAR data is transformed to the RTK ground-truth coordinate by the calibration parameters provided in the dataset, so that the ground-truth of the transformation from the initial source point cloud to the initial target point cloud can be considered as a unit matrix. The input target point cloud $P_t$ of each set of experiments is the initial target point cloud $P_t^i$, and the input source point cloud $P_s$ is obtained from the initial source point cloud through a random transformation $T_{rm}$, the rotation range of this random transformation is \SIrange{0}{30}{\degree}, and the translation range is \SIrange{0}{40}{\meter}. Each method under each set of sequences does 50 repetitions of experiments, and takes the average of the evaluation metric to make comparisons. All tests were done on a desktop computer with 32GB RAM, a 12GB RTX3060 graphics card and an  Intel (R) Core (TM) i5-13500 CPU.
The initial target point cloud $\mathcal{P}_{\text{target}}^{\text{i}}$ and source point cloud $\mathcal{P}_{\text{source}}^{\text{i}}$ are obtained from the LiDAR raw data. Each frame of LiDAR data is transformed to the RTK ground-truth coordinate using the calibration parameters provided by GrAco. Consequently, the ground-truth of the transformation from $\mathcal{P}_{\text{source}}^{\text{i}}$ to $\mathcal{P}_{\text{target}}^{\text{i}}$ can be considered as a unit matrix. The input target point cloud $\mathcal{P}_{\text{source}}$ for each experiment set is derived from $\mathcal{P}_{\text{target}}^{\text{i}}$, while the input source point cloud $\mathcal{P}_{\text{source}}$ is obtained from $\mathcal{P}_{\text{source}}^{\text{i}}$ through a random transformation $T_{rm}$. The random transformation possesses a rotation range of \SIrange{0}{90}{\degree} and a translation range of \SIrange{0}{100}{\meter}. Each method within each combination undergoes 100 repetitions of experiments, and the evaluation metric average is computed for comparative analysis. All experiments were performed on a desktop computer equipped with 32GB RAM, a 12GB RTX3060 graphics card.
%and an  Intel (R) Core (TM) i5-13500 CPU.

\subsection{Comparisons}

% The performance of the proposed method is compared with four point cloud registration methods using classical point cloud feature descriptors, including FPFH+TEASER, FPFH+SVD, SHOT+SVD, and SpinImage+SVD, in which FPFH+TEASER is implemented using the open3d library in python, and the last three methods are implemented using the PCL library in C++. The rotation error, translation error, RMSD and registration success rate of each method in each sequence combination are presented in Table \ref{table:rot_trans_error} and Table \ref{table:srr}, respectively. Moreover, The results of the registration of each method are shown in Fig\ref{fig:result}.
% \input{table/ssr}

The performance of \pcr is evaluated in comparison with four point cloud registration methods, namely FPFH+TEASER\cite{yang2020teaser}, PLADE\cite{chen2019plade}, Predator\cite{huang2021predator} and GeoTransformer\cite{qin2023geotransformer}. The first two methods rely on hand-crafted descriptors, while the latter two are based on deep learning networks. Rotation error, translation error, root mean square distance (RMSD), and registration success rates for each method across various sequence combinations are documented in Table.\ref{table:rot_trans_error} and Table.\ref{table:srr}, respectively. Furthermore, the outcomes of the registration for each method are visually depicted in Fig.\ref{fig:result}.
\input{table/ssr}

The results from Table.\ref{table:rot_trans_error} suggest that \pcr exhibits superior performance compared to other methods across most sequence combinations, achieving the lowest rotation error, translation error, and RSMD. Notably, the method's exceptional performance is observed in combinations such as A03-G05 and A04-G06. 
% For instance, in the A03-G05 combination, the proposed method demonstrates a rotation error of \SI{0.414}{\degree}, a translation error of \SI{0.794}{\meter}, and an RMSD of \SI{0.644}{\meter} across an expansive area. 
In contrast, the errors of other methods fluctuate between tens and hundreds of degrees and meters.
The observed subpar performance can be mainly attributed to the limited overlap and the dearth of local geometric features present in the point clouds. It is evident that the performance of these methods is enhanced when the scans exhibit significantly higher overlap, e.g., the A-A and G-G combinations shown in Table.\ref{table:dataset}.
From Table.\ref{table:srr}, it is clear that \pcr outperforms other methods in all scenarios with success rates of 97.25\%, 96.67\%, and 96.50\% respectively. This signifies that \pcr demonstrates greater robustness and higher success rates across various point cloud registration scenarios.

% Based on the overall results in the table, \pcr demonstrates strong performance across all types of point cloud combinations. It maintains low errors in complex A-G combinations as well as in relatively simple A-A and G-G combinations. This indicates that the method exhibits high robustness and generality when dealing with large-scale point cloud registration with limited overlap.
Based on the overall results in the table, \pcr demonstrates strong performance across all point cloud combinations. It maintains low errors in complex A-G combinations as well as in A-A and G-G combinations, indicating high robustness and generality in large-scale point cloud registration with limited overlap.

\subsection{Ablation Study}
% \enote{Need to be modified}
\begin{figure}[t]
	\centering
	\includegraphics[width = \linewidth]{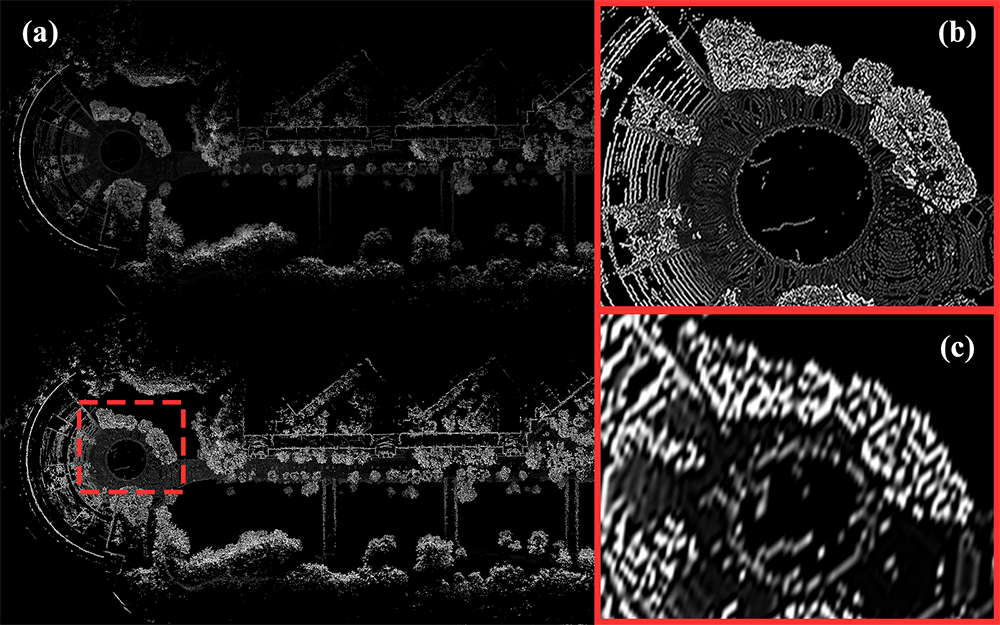}
	\caption{Effects of using different modules in the ablation experiments: (a) BEV image of G03 without image enhancement is located at the top, and the enhanced image is at the bottom.; (b) and (c) are parts with and without resolution augmentation in the red dashed box in (a), respectively}
	\label{fig:ablation}
\end{figure}

\input{table/ablation}

% Furthermore, an in-depth performance evaluation of the adopted modules for enhancing the point cloud registration performance validates the effectiveness of the method, and the results are shown in Table. \ref{table:ablation}. Compared to the proposed method, the removal of the three modules leads to a decrease in the registration performance or even a failure of the registration. By increasing the resolution of the map (RES for short), feature points that would otherwise overlap can be distinguished, substantially reducing the error of the registration, as shown in Figure. \ref{fig:ablation}(b)\&(c). By enhancing the BEV image (IE for short) as shown in Figure \ref{fig:ablation}.(a), the structures in the scene can be better differentiated, which greatly improves the registration performance by more than 10 times. In addition, by adding the FOCUS module, matching is done again in the area where feature points are concentrated, which has a performance improvement for the repositioning of small maps in large area, as shown in Figure. \ref{fig:focus}. The combination of all three components leads to the best overall performance.

In addition, a comprehensive performance evaluation of the modules used to improve the point cloud registration performance confirms the method's effectiveness. Detailed results are presented in Table.\ref{table:ablation}. Removal of the three modules resulted in a noticeable decrease in registration performance, and in some cases, even failure to register. 
Adjusting the resolution of the point clouds (RES) enables the distinction of keypoints that would otherwise overlap, significantly reducing registration error, as illustrated in Fig.\ref{fig:ablation}(b)\&(c). 
Enhancing the BEV image(IE), as depicted in Fig.\ref{fig:ablation}(a), facilitates improved differentiation of scene structures, culminating in a more than tenfold improvement in registration performance. 
Furthermore, the incorporation of the FOCUS module, facilitating re-matching in keypoint-dense areas, notably enhances the repositioning of limited overlap within larger areas, as depicted in Fig.\ref{fig:focus}. The collective integration of all three components yields the most superior overall performance.

%% file: table/dataset.tex
\begin{table}[t]
\center
\caption{Scale and Overlap Information of Typical Combination in the GrAco Dataset. }
\label{table:dataset}
\renewcommand{\arraystretch}{1.5}
{
\resizebox{\linewidth}{!}{
\begin{tabular}{ccccc}
\hline
\multirow{2}{*}{Source} & \multirow{2}{*}{Target} & \multirow{2}{*}{Scene Scale}                                                  & \multicolumn{2}{c}{Overlap Ratio} \\ \cline{4-5} 
                        &                         &                                                                        & Source          & Target          \\ \hline
A03                     & G03                     & $\SI{440.21}{\meter}\times\SI{454.04}{\meter}\times\SI{58.58}{\meter}$ & 14.82\%         & 22.28\%         \\
A04                     & G01                     & $\SI{457.13}{\meter}\times\SI{387.83}{\meter}\times\SI{32.91}{\meter}$ & 19.97\%         & 22.90\%         \\
A07                     & G06                     & $\SI{432.11}{\meter}\times\SI{492.93}{\meter}\times\SI{33.35}{\meter}$ & 16.09\%         & 18.10\%         \\
G04                     & G05                     & $\SI{558.95}{\meter}\times\SI{466.31}{\meter}\times\SI{26.72}{\meter}$ & 66.49\%         & 42.11\%         \\
A01                     & A02                     & $\SI{390.39}{\meter}\times\SI{452.18}{\meter}\times\SI{57.55}{\meter}$ & 32.42\%         & 75.19\%         \\ \hline
\end{tabular}}
}

\renewcommand{\arraystretch}{1}
\end{table}

%% file: table/rot_trans_error.tex
\begin{table*}[ht]
\center
\caption{average errors in all point cloud combinations of different methods}
\label{table:rot_trans_error}
\renewcommand{\arraystretch}{1.5}
\resizebox{\linewidth}{!}{

\begin{tabular}{c|l|cccccccc|ccc|ccc}
\hline
\multirow{2}{*}{Methods}                                                   & \multicolumn{1}{c|}{Source}                 & A02            & A03            & A03            & A04            & A04            & A06            & A07            & A07            & G01            & G03            & G04            & A01            & A05            & A06            \\
                                                                           & \multicolumn{1}{c|}{Target}                 & G05            & G03            & G05            & G01            & G06            & G01            & G05            & G06            & G02            & G06            & G05            & A02            & A07            & A08            \\ \hline\hline
\multirow{3}{*}{\begin{tabular}[c]{@{}c@{}}FPFH+\\ TEASER\end{tabular}}    & $e_{s, t}^{r}$($^\circ$)$\downarrow$        & 1.927          & 2.262          & 13.494         & 116.870        & 135.276        & 2.258          & 19.538         & 18.782         & 0.961          & 1.935          & 0.330          & 1.946          & 0.354          & \textbf{0.536} \\
                                                                           & $e_{s, t}^{t}$($\SI{}{\meter}$)$\downarrow$ & 7.430          & 2.351          & 20.053         & 122.073        & 133.817        & 4.141          & 37.906         & 20.387         & 1.444          & 2.163          & 1.062          & 8.619          & \textbf{0.746} & \textbf{0.646} \\
                                                                           & $RMSD$($\SI{}{\meter}$)$\downarrow$         & 1.472          & 1.797          & 13.718         & 156.847        & 171.099        & 2.368          & 34.720         & 16.062         & 1.128          & 3.743          & 0.498          & 2.314          & \textbf{0.912} & 0.735          \\ \hline
% \multirow{3}{*}{FPFH+SVD}                                                  & $e_{s, t}^{r}$($^\circ$)$\downarrow$        & 5.120          & 13.064         & 13.590         & 148.059        & 83.700         & 14.278         & 6.152          & 7.096          & 12.422         & 17.946         & 2.828          & 8.294          & 5.077          & 12.483         \\
%                                                                            & $e_{s, t}^{t}$($\SI{}{\meter}$)$\downarrow$ & 55.235         & 6.711          & 21.200         & 80.520         & 74.290         & 62.914         & 126.617        & 19.251         & 72.176         & 105.874        & 11.906         & 40.512         & 21.774         & 28.006         \\
%                                                                            & $RMSD$($\SI{}{\meter}$)$\downarrow$         & 59.234         & 10.355         & 32.425         & 89.095         & 93.452         & 70.004         & 132.943        & 19.876         & 77.030         & 102.462        & 9.782          & 38.344         & 18.455         & 29.984         \\ \hline
\multirow{3}{*}{PLADE}                                                     & $e_{s, t}^{r}$($^\circ$)$\downarrow$        & 39.621         & 28.063         & 6.376          & 48.336         & 15.019         & 3.954          & 8.215          & 44.288         & 9.579          & 1.058          & 5.489          & 16.870         & 29.966         & 6.832          \\
                                                                           & $e_{s, t}^{t}$($\SI{}{\meter}$)$\downarrow$ & 105.302        & 36.386         & 60.531         & 46.213         & 7.067          & 22.976         & 10.891         & 47.040         & 8.642          & 5.582          & 8.801          & 130.142        & 125.632        & 106.821        \\
                                                                           & $RMSD$($\SI{}{\meter}$)$\downarrow$         & 112.395        & 33.748         & 52.759         & 37.947         & 7.274          & 24.189         & 13.608         & 48.299         & 9.993          & 5.914          & 7.668          & 123.813        & 133.473        & 109.133        \\ \hline
% \multirow{3}{*}{\begin{tabular}[c]{@{}l@{}}SpinImage+\\ SVD\end{tabular}}  & $e_{s, t}^{r}$($^\circ$)$\downarrow$        & 85.024         & 149.601        & 32.379         & 127.438        & 24.666         & 32.634         & 99.331         & 52.581         & 179.142        & 1.335          & 6.251          & 16.010         & 24.732         & 14.331         \\
%                                                                            & $e_{s, t}^{t}$($\SI{}{\meter}$)$\downarrow$ & 211.419        & 129.440        & 83.437         & 83.829         & 138.882        & 62.072         & 246.243        & 70.229         & 51.670         & 84.605         & 54.349         & 28.902         & 85.737         & 70.343         \\
%                                                                            & $RMSD$($\SI{}{\meter}$)$\downarrow$         & 200.246        & 136.747        & 90.904         & 87.622         & 144.623        & 65.976         & 263.335        & 77.647         & 57.186         & 89.912         & 55.489         & 32.341         & 88.333         & 76.037         \\ \hline
\multirow{3}{*}{Predator}                                                  & $e_{s, t}^{r}$($^\circ$)$\downarrow$        & 6.377          & 7.193          & 23.590         & 20.208         & 14.112         & 10.242         & 3.292          & 7.888          & 1.115          & 1.133          & 1.015          & 4.125          & 1.389          & 2.242          \\
                                                                           & $e_{s, t}^{t}$($\SI{}{\meter}$)$\downarrow$ & 4.782          & 10.465         & 21.200         & 68.037         & 31.710         & 5.588          & 18.964         & 6.415          & 1.143          & 1.690          & 1.820          & 6.186          & 1.830          & 1.970          \\
                                                                           & $RMSD$($\SI{}{\meter}$)$\downarrow$         & 2.805          & 11.403         & 32.425         & 60.327         & 38.990         & 3.151          & 20.726         & 6.921          & 1.327          & 1.840          & 1.749          & 5.379          & 1.466          & 1.379          \\ \hline
\multirow{3}{*}{GeoTransformer}                                            & $e_{s, t}^{r}$($^\circ$)$\downarrow$        & 1.901          & 5.756          & 12.026         & 14.928         & 45.002         & 3.385          & 10.246         & 9.189          & 1.310          & \textbf{0.708} & 0.297          & 8.167          & 0.384          & 1.558          \\
                                                                           & $e_{s, t}^{t}$($\SI{}{\meter}$)$\downarrow$ & 8.455          & 4.537          & 5.120          & 10.211         & 25.512         & 3.944          & 6.598          & 5.479          & \textbf{0.715} & 1.346          & 1.144          & 4.889          & 0.918          & 1.124          \\
                                                                           & $RMSD$($\SI{}{\meter}$)$\downarrow$         & 6.490          & 4.626          & 4.381          & 12.196         & 20.839         & 1.674          & 10.723         & 6.305          & 0.767          & 1.708          & 1.531          & 3.036          & 1.002          & 1.383          \\ \hline
\multirow{3}{*}{\begin{tabular}[c]{@{}c@{}}\pcr\\ (Ours)\end{tabular}} & $e_{s, t}^{r}$($^\circ$)$\downarrow$        & \textbf{0.320} & \textbf{0.504} & \textbf{0.414} & \textbf{2.825} & \textbf{0.892} & \textbf{0.776} & \textbf{0.862} & \textbf{0.511} & \textbf{0.400} & 0.731          & \textbf{0.133} & \textbf{0.586} & \textbf{0.240} & {0.575} \\
                                                                           & $e_{s, t}^{t}$($\SI{}{\meter}$)$\downarrow$ & \textbf{1.156} & \textbf{1.278} & \textbf{0.794} & \textbf{3.021} & \textbf{1.212} & \textbf{1.567} & \textbf{0.827} & \textbf{1.010} & 0.820          & \textbf{0.825} & \textbf{0.395} & \textbf{1.750} & {0.914} & 1.007 \\
                                                                           & $RMSD$($\SI{}{\meter}$)$\downarrow$         & \textbf{0.372} & \textbf{1.750} & \textbf{0.644} & \textbf{4.128} & \textbf{1.616} & \textbf{0.882} & \textbf{0.981} & \textbf{1.222} & \textbf{0.490} & \textbf{1.671} & \textbf{0.271} & \textbf{0.914} & {0.979} & \textbf{0.713} \\ \hline
\end{tabular}}

\renewcommand{\arraystretch}{1}
\end{table*}

%% file: table/ssr.tex
\begin{table}[t]
\center
\caption{successful registration rate of different methods}
\label{table:srr}
\renewcommand{\arraystretch}{1.5}
{
\begin{tabular}{c|ccc}
\hline
Method          & A-G    & G-G    & A-A    \\ \hline
FPFH+TEASER     & 36.88\% & 86.25\% & 65.00\% \\
% FPFH+SVD        & 2.19\%  & 6.25\%  & 1.50\%  \\
PLADE           & 7.48\%  & 44.75\% & 5.25\%  \\
% SpinImage+SVD   & 0.00\%  & 2.08\%  & 0.00\%  \\
Predator        & 40.44\% & 81.50\% & 77.32\%  \\
GeoTransformer  & 63.72\% & 92.48\% & 87.64\%  \\
\pcr(Ours) & 97.25\% & 96.67\% & 96.50\% \\ \hline
\end{tabular}
}

\renewcommand{\arraystretch}{1}
\end{table}

%% file: table/ablation.tex
% \usepackage{multirow}
\begin{table}[t]
\center
\caption{abaltion study}
\label{table:ablation}
\renewcommand{\arraystretch}{1.5}
\resizebox{\linewidth}{!}{
\begin{tabular}{ccc|ccc|ccc}
\hline
\multicolumn{3}{c|}{registration   model} & \multicolumn{3}{c|}{A03-G05}                     & \multicolumn{3}{c}{A06-G01}                      \\ \hline
RES    & FOCUS       & IE    & $e_r$            & $e_t$            & RMSD           & $e_r$            & $e_t$            & RMSD           \\ \hline
\ding{56}     & \ding{56}   & \ding{56}   & 8.033          & 10.237         & 9.604          & 8.408          & 20.537         & 20.326         \\
\ding{51}     & \ding{51}   & \ding{51}   & \textbf{0.414} & \textbf{0.794} & \textbf{0.644} & \textbf{0.776} & \textbf{0.567} & \textbf{0.882} \\
\ding{56}     & \ding{51}   & \ding{51}   & 1.865          & 1.738          & 0.989          & 3.869          & 18.502         & 16.792         \\
\ding{51}     & \ding{56}   & \ding{51}   & 3.083          & 3.541          & 2.083          & 1.941          & 9.217          & 8.712          \\
\ding{51}     & \ding{51}   & \ding{56}   & 9.009          & 7.940          & 6.932          & 4.350          & 18.606         & 16.339         \\ \hline
\end{tabular}
}
\renewcommand{\arraystretch}{1}
\end{table}

%% file: chapters/conclusion.tex
\section{Conclusion}

% In this paper, an accurate, stable and robust ground- aerial LiDAR point cloud registration algorithm named SuperICP is proposed. The proposed method utilizes the idea of image feature point matching to solve the problem of point cloud correspondence estimation. The proposed method has significant advantages over the commonly used feature-matching-based point cloud registration algorithms in eight sets of ground-aerial sequence combinations with low overlap and squared metric level . In future research, the robustness and efficiency of the algorithm can continue to be improved. For this purpose, some methods such as using BEV maps of LiDAR point clouds specifically train an image feature recognition network. Furthermore, the performance of the proposed method can be evaluated on data from different sensors.

In this paper, we propose \pcr for the registration of large-scale scene point clouds with limited overlap, focusing specifically on TLS-ALS registration. Based on modality transformation, \pcr converts 3D point clouds into 2D BEV images, facilitating correspondence estimation through 2D image keypoints matching. Our method has advantages in ground-aerial collaborative reconstruction scenarios over commonly utilized registration algorithms. In future work, \pcr can be integrated into multi-robot collaborative systems and explored for applications involving the fusion of multimodal data.

%% file: chapters/acknowledgments.tex
\section*{Acknowledgment} This work is supported by the National Natural Science Foundation of China (No. 62332016) and the Chinese Academy of Sciences Frontier Science Key Research Project ZDBS-LY-JSC001.

%% file: root.bbl
\begin{thebibliography}{10}
\providecommand{\url}[1]{#1}
\csname url@samestyle\endcsname
\providecommand{\newblock}{\relax}
\providecommand{\bibinfo}[2]{#2}
\providecommand{\BIBentrySTDinterwordspacing}{\spaceskip=0pt\relax}
\providecommand{\BIBentryALTinterwordstretchfactor}{4}
\providecommand{\BIBentryALTinterwordspacing}{\spaceskip=\fontdimen2\font plus
\BIBentryALTinterwordstretchfactor\fontdimen3\font minus \fontdimen4\font\relax}
\providecommand{\BIBforeignlanguage}[2]{{%
\expandafter\ifx\csname l@#1\endcsname\relax
\typeout{** WARNING: IEEEtran.bst: No hyphenation pattern has been}%
\typeout{** loaded for the language `#1'. Using the pattern for}%
\typeout{** the default language instead.}%
\else
\language=\csname l@#1\endcsname
\fi
#2}}
\providecommand{\BIBdecl}{\relax}
\BIBdecl

\bibitem{pomerleau2015review}
F.~Pomerleau, F.~Colas, R.~Siegwart \emph{et~al.}, ``A review of point cloud registration algorithms for mobile robotics,'' \emph{Foundations and Trends{\textregistered} in Robotics}, vol.~4, no.~1, pp. 1--104, 2015.

\bibitem{nagy2018real}
B.~Nagy and C.~Benedek, ``Real-time point cloud alignment for vehicle localization in a high resolution 3d map,'' in \emph{Proceedings of the european conference on computer vision (ECCV) workshops}, 2018, pp. 0--0.

\bibitem{dong2018hierarchical}
Z.~Dong, B.~Yang, F.~Liang, R.~Huang, and S.~Scherer, ``Hierarchical registration of unordered tls point clouds based on binary shape context descriptor,'' \emph{ISPRS Journal of Photogrammetry and Remote Sensing}, vol. 144, pp. 61--79, 2018.

\bibitem{ghorbani2022novel}
F.~Ghorbani, H.~Ebadi, A.~Sedaghat, and N.~Pfeifer, ``A novel 3-d local daisy-style descriptor to reduce the effect of point displacement error in point cloud registration,'' \emph{IEEE Journal of Selected Topics in Applied Earth Observations and Remote Sensing}, vol.~15, pp. 2254--2273, 2022.

\bibitem{lu2021hregnet}
F.~Lu, G.~Chen, Y.~Liu, L.~Zhang, S.~Qu, S.~Liu, and R.~Gu, ``Hregnet: A hierarchical network for large-scale outdoor lidar point cloud registration,'' in \emph{Proceedings of the IEEE/CVF International Conference on Computer Vision}, 2021, pp. 16\,014--16\,023.

\bibitem{yu2023peal}
J.~Yu, L.~Ren, Y.~Zhang, W.~Zhou, L.~Lin, and G.~Dai, ``Peal: Prior-embedded explicit attention learning for low-overlap point cloud registration,'' in \emph{Proceedings of the IEEE/CVF Conference on Computer Vision and Pattern Recognition}, 2023, pp. 17\,702--17\,711.

\bibitem{zeng20173dmatch}
A.~Zeng, S.~Song, M.~Nie{\ss}ner, M.~Fisher, J.~Xiao, and T.~Funkhouser, ``3dmatch: Learning local geometric descriptors from rgb-d reconstructions,'' in \emph{Proceedings of the IEEE conference on computer vision and pattern recognition}, 2017, pp. 1802--1811.

\bibitem{ao2021spinnet}
S.~Ao, Q.~Hu, B.~Yang, A.~Markham, and Y.~Guo, ``Spinnet: Learning a general surface descriptor for 3d point cloud registration,'' in \emph{Proceedings of the IEEE/CVF conference on computer vision and pattern recognition}, 2021, pp. 11\,753--11\,762.

\bibitem{detone2018superpoint}
D.~DeTone, T.~Malisiewicz, and A.~Rabinovich, ``Superpoint: Self-supervised interest point detection and description,'' in \emph{Proceedings of the IEEE conference on computer vision and pattern recognition workshops}, 2018, pp. 224--236.

\bibitem{lindenberger2023lightglue}
P.~Lindenberger, P.-E. Sarlin, and M.~Pollefeys, ``Lightglue: Local feature matching at light speed,'' in \emph{Proceedings of the IEEE/CVF International Conference on Computer Vision}, 2023, pp. 17\,627--17\,638.

\bibitem{prokop2019low}
M.~Prokop, S.~A. Shaikh, and K.-S. Kim, ``Low overlapping point cloud registration using line features detection,'' \emph{Remote Sensing}, vol.~12, no.~1, p.~61, 2019.

\bibitem{dong2023application}
Y.~Dong, B.~Xu, T.~Liao, C.~Yin, and Z.~Tan, ``Application of local-feature-based 3d point cloud stitching method of low-overlap point cloud to aero-engine blade measurement,'' \emph{IEEE Transactions On Instrumentation And Measurement}, 2023.

\bibitem{lei2017fast}
H.~Lei, G.~Jiang, and L.~Quan, ``Fast descriptors and correspondence propagation for robust global point cloud registration,'' \emph{IEEE Transactions on Image Processing}, vol.~26, no.~8, pp. 3614--3623, 2017.

\bibitem{xu2019pairwise}
Y.~Xu, R.~Boerner, W.~Yao, L.~Hoegner, and U.~Stilla, ``Pairwise coarse registration of point clouds in urban scenes using voxel-based 4-planes congruent sets,'' \emph{ISPRS journal of photogrammetry and remote sensing}, vol. 151, pp. 106--123, 2019.

\bibitem{li2021automatic}
J.~Li, S.~Huang, H.~Cui, Y.~Ma, and X.~Chen, ``Automatic point cloud registration for large outdoor scenes using a priori semantic information,'' \emph{Remote Sensing}, vol.~13, no.~17, p. 3474, 2021.

\bibitem{liu2021coarse}
W.~Liu, W.~Sun, S.~Wang, and Y.~Liu, ``Coarse registration of point clouds with low overlap rate on feature regions,'' \emph{Signal Processing: Image Communication}, vol.~98, p. 116428, 2021.

\bibitem{duan2024rotation}
Y.~Duan, X.~Zhang, G.~You, Y.~Wu, X.~Li, Y.~Li, X.~Chu, J.~Peng, Y.~Zhang, J.~Ji \emph{et~al.}, ``Rotation initialization and stepwise refinement for universal lidar calibration,'' \emph{arXiv preprint arXiv:2405.05589}, 2024.

\bibitem{chen2019plade}
S.~Chen, L.~Nan, R.~Xia, J.~Zhao, and P.~Wonka, ``Plade: A plane-based descriptor for point cloud registration with small overlap,'' \emph{IEEE Transactions on Geoscience and Remote Sensing}, vol.~58, no.~4, pp. 2530--2540, 2019.

\bibitem{huang2021predator}
S.~Huang, Z.~Gojcic, M.~Usvyatsov, A.~Wieser, and K.~Schindler, ``Predator: Registration of 3d point clouds with low overlap,'' in \emph{Proceedings of the IEEE/CVF Conference on computer vision and pattern recognition}, 2021, pp. 4267--4276.

\bibitem{gao2019ground}
X.~Gao, S.~Shen, Z.~Hu, and Z.~Wang, ``Ground and aerial meta-data integration for localization and reconstruction: A review,'' \emph{Pattern Recognition Letters}, vol. 127, pp. 202--214, 2019.

\bibitem{zhu2023graco}
Y.~Zhu, Y.~Kong, Y.~Jie, S.~Xu, and H.~Cheng, ``Graco: A multimodal dataset for ground and aerial cooperative localization and mapping,'' \emph{IEEE Robotics and Automation Letters}, vol.~8, no.~2, pp. 966--973, 2023.

\bibitem{he2020ground}
J.~He, Y.~Zhou, L.~Huang, Y.~Kong, and H.~Cheng, ``Ground and aerial collaborative mapping in urban environments,'' \emph{IEEE Robotics and Automation Letters}, vol.~6, no.~1, pp. 95--102, 2020.

\bibitem{li2021tutorial}
L.~Li, R.~Wang, and X.~Zhang, ``A tutorial review on point cloud registrations: principle, classification, comparison, and technology challenges,'' \emph{Mathematical Problems in Engineering}, vol. 2021, no.~1, p. 9953910, 2021.

\bibitem{mallot1991inverse}
H.~A. Mallot, H.~H. B{\"u}lthoff, J.~J. Little, and S.~Bohrer, ``Inverse perspective mapping simplifies optical flow computation and obstacle detection,'' \emph{Biological cybernetics}, vol.~64, no.~3, pp. 177--185, 1991.

\bibitem{besl1992method}
P.~J. Besl and N.~D. McKay, ``Method for registration of 3-d shapes,'' in \emph{Sensor fusion IV: control paradigms and data structures}, vol. 1611.\hskip 1em plus 0.5em minus 0.4em\relax Spie, 1992, pp. 586--606.

\bibitem{rueckert1999nonrigid}
D.~Rueckert, L.~I. Sonoda, C.~Hayes, D.~L. Hill, M.~O. Leach, and D.~J. Hawkes, ``Nonrigid registration using free-form deformations: application to breast mr images,'' \emph{IEEE transactions on medical imaging}, vol.~18, no.~8, pp. 712--721, 1999.

\bibitem{rusinkiewicz2001efficient}
S.~Rusinkiewicz and M.~Levoy, ``Efficient variants of the icp algorithm,'' in \emph{Proceedings third international conference on 3-D digital imaging and modeling}.\hskip 1em plus 0.5em minus 0.4em\relax IEEE, 2001, pp. 145--152.

\bibitem{li2008global}
H.~Li, R.~W. Sumner, and M.~Pauly, ``Global correspondence optimization for non-rigid registration of depth scans,'' in \emph{Computer graphics forum}, vol.~27, no.~5.\hskip 1em plus 0.5em minus 0.4em\relax Wiley Online Library, 2008, pp. 1421--1430.

\bibitem{segal2009generalized}
A.~Segal, D.~Haehnel, and S.~Thrun, ``Generalized-icp.'' in \emph{Robotics: science and systems}, vol.~2, no.~4.\hskip 1em plus 0.5em minus 0.4em\relax Seattle, WA, 2009, p. 435.

\bibitem{gelfand2005robust}
N.~Gelfand, N.~J. Mitra, L.~J. Guibas, and H.~Pottmann, ``Robust global registration,'' in \emph{Symposium on geometry processing}, vol.~2, no.~3.\hskip 1em plus 0.5em minus 0.4em\relax Vienna, Austria, 2005, p.~5.

\bibitem{qi2017pointnet}
C.~R. Qi, H.~Su, K.~Mo, and L.~J. Guibas, ``Pointnet: Deep learning on point sets for 3d classification and segmentation,'' in \emph{Proceedings of the IEEE conference on computer vision and pattern recognition}, 2017, pp. 652--660.

\bibitem{elbaz20173d}
G.~Elbaz, T.~Avraham, and A.~Fischer, ``3d point cloud registration for localization using a deep neural network auto-encoder,'' in \emph{Proceedings of the IEEE conference on computer vision and pattern recognition}, 2017, pp. 4631--4640.

\bibitem{deng2018ppfnet}
H.~Deng, T.~Birdal, and S.~Ilic, ``Ppfnet: Global context aware local features for robust 3d point matching,'' in \emph{Proceedings of the IEEE conference on computer vision and pattern recognition}, 2018, pp. 195--205.

\bibitem{choy2019fully}
C.~Choy, J.~Park, and V.~Koltun, ``Fully convolutional geometric features,'' in \emph{Proceedings of the IEEE/CVF international conference on computer vision}, 2019, pp. 8958--8966.

\bibitem{rusu2009fast}
R.~B. Rusu, N.~Blodow, and M.~Beetz, ``Fast point feature histograms (fpfh) for 3d registration,'' in \emph{2009 IEEE international conference on robotics and automation}.\hskip 1em plus 0.5em minus 0.4em\relax IEEE, 2009, pp. 3212--3217.

\bibitem{tombari2010unique}
F.~Tombari, S.~Salti, and L.~Di~Stefano, ``Unique signatures of histograms for local surface description,'' in \emph{Computer Vision--ECCV 2010: 11th European Conference on Computer Vision, Heraklion, Crete, Greece, September 5-11, 2010, Proceedings, Part III 11}.\hskip 1em plus 0.5em minus 0.4em\relax Springer, 2010, pp. 356--369.

\bibitem{wolfson1997geometric}
H.~J. Wolfson and I.~Rigoutsos, ``Geometric hashing: An overview,'' \emph{IEEE computational science and engineering}, vol.~4, no.~4, pp. 10--21, 1997.

\bibitem{fischler1981random}
M.~A. Fischler and R.~C. Bolles, ``Random sample consensus: a paradigm for model fitting with applications to image analysis and automated cartography,'' \emph{Communications of the ACM}, vol.~24, no.~6, pp. 381--395, 1981.

\bibitem{chen1999ransac}
C.-S. Chen, Y.-P. Hung, and J.-B. Cheng, ``Ransac-based darces: A new approach to fast automatic registration of partially overlapping range images,'' \emph{IEEE Transactions on Pattern Analysis and Machine Intelligence}, vol.~21, no.~11, pp. 1229--1234, 1999.

\bibitem{yang2020teaser}
H.~Yang, J.~Shi, and L.~Carlone, ``Teaser: Fast and certifiable point cloud registration,'' \emph{IEEE Transactions on Robotics}, vol.~37, no.~2, pp. 314--333, 2020.

\bibitem{ren2022corri2p}
S.~Ren, Y.~Zeng, J.~Hou, and X.~Chen, ``Corri2p: Deep image-to-point cloud registration via dense correspondence,'' \emph{IEEE Transactions on Circuits and Systems for Video Technology}, vol.~33, no.~3, pp. 1198--1208, 2022.

\bibitem{zhang2022pcr}
Y.~Zhang, J.~Yu, X.~Huang, W.~Zhou, and J.~Hou, ``Pcr-cg: Point cloud registration via deep explicit color and geometry,'' in \emph{European Conference on Computer Vision}.\hskip 1em plus 0.5em minus 0.4em\relax Springer, 2022, pp. 443--459.

\bibitem{chen2022imlovenet}
H.~Chen, Z.~Wei, Y.~Xu, M.~Wei, and J.~Wang, ``Imlovenet: Misaligned image-supported registration network for low-overlap point cloud pairs,'' in \emph{ACM SIGGRAPH 2022 conference proceedings}, 2022, pp. 1--9.

\bibitem{huang2022imfnet}
X.~Huang, W.~Qu, Y.~Zuo, Y.~Fang, and X.~Zhao, ``Imfnet: Interpretable multimodal fusion for point cloud registration,'' \emph{IEEE Robotics and Automation Letters}, vol.~7, no.~4, pp. 12\,323--12\,330, 2022.

\bibitem{gonzalez2009digital}
R.~C. Gonzalez, \emph{Digital image processing}.\hskip 1em plus 0.5em minus 0.4em\relax Pearson education india, 2009.

\bibitem{golub1971singular}
G.~H. Golub and C.~Reinsch, ``Singular value decomposition and least squares solutions,'' in \emph{Handbook for Automatic Computation: Volume II: Linear Algebra}.\hskip 1em plus 0.5em minus 0.4em\relax Springer, 1971, pp. 134--151.

\bibitem{qin2023geotransformer}
Z.~Qin, H.~Yu, C.~Wang, Y.~Guo, Y.~Peng, S.~Ilic, D.~Hu, and K.~Xu, ``Geotransformer: Fast and robust point cloud registration with geometric transformer,'' \emph{IEEE Transactions on Pattern Analysis and Machine Intelligence}, vol.~45, no.~8, pp. 9806--9821, 2023.

\end{thebibliography}
